\documentclass{llncs}
\usepackage{./llncsdoc}
\usepackage{amsmath}
\usepackage{amssymb}
\usepackage{caption}
\usepackage{cite}
\usepackage{bm}
\usepackage{mathrsfs}
\usepackage{graphicx}
\usepackage{epsfig}
\usepackage{pdfpages}
\usepackage{wrapfig}
\usepackage{lipsum}
\usepackage{times}
\usepackage{setspace}
\usepackage{cancel} 
\usepackage{color}
\usepackage{wrapfig}
\usepackage{adjustbox}
\usepackage{xspace}
\usepackage{float}
\usepackage{cases}
\usepackage[colorlinks, citecolor=black, linkcolor=blue, linktocpage=true]{hyperref}
\usepackage[retainorgcmds]{IEEEtrantools}
\usepackage[hang]{subfigure}
\newcommand{\ignore}[1]{}

\newcommand{\mbfdot}[1]{\dot{\mbf{#1}}}

\newcommand{\norm}[1]{\left\Vert#1\right\Vert} 
\newcommand{\abs}[1]{\left\vert#1\right\vert} 
\newcommand{\bbm}{\begin{bmatrix}}
\newcommand{\ebm}{\end{bmatrix}}
\newcommand{\bma}[1]{\left[\begin{array}{#1}}
\newcommand{\ema}{\end{array}\right]}
\DeclareMathAlphabet{\mbf}{OT1}{ptm}{b}{n}
\newcommand{\mbs}[1]{{\boldsymbol{#1}}}
 
\newcommand{\mbfbar}[1]{{\bar{\mbf{#1}}}}

\def\dotb{{\raisebox{-0.6ex}{ \kern0.2ex\raisebox{0.8ex}{\tiny $\circ$}}}}
\def\ddota{{\raisebox{-0.6ex}{ \raise0.2ex\hbox{ \LARGE $\cdot\hspace*{-0.2ex}\cdot$}}}}
\def\ddotb{{\raisebox{-0.6ex}{ \kern0.2ex\raisebox{0.8ex}{\tiny $\circ\circ$}}}}
\newcommand{\p}{\partial}

\newcommand{\trans}{{\ensuremath{\mathsf{T}}}} 
\newcommand{\beq}{\begin{equation}}
\newcommand{\eeq}{\end{equation}}
\newcommand{\bdis}{\begin{displaymath}}
\newcommand{\edis}{\end{displaymath}}
\newcommand{\beqarray}{\begin{eqnarray}}
\newcommand{\eeqarray}{\end{eqnarray}}
\newcommand{\beqarraynn}{\begin{eqnarray*}}
\newcommand{\eeqarraynn}{\end{eqnarray*}}
\newcommand{\ora}[1]{\overrightarrow{#1}}
\title{\LARGE \bf
Feedback Control of the Pusher-Slider System: A Story of Hybrid and Underactuated \\Contact Dynamics}

\author{Fran\c{c}ois Robert Hogan  \hspace{.1mm} and  \hspace{.5mm}
Alberto Rodriguez
}
  \institute{Department of Mechanical Engineering, Massachussetts Institute of Technology,\\ 77 Massachusetts Avenue, Cambridge, MA, USA}
\begin{document}
\maketitle
\begin{abstract}
  This paper investigates real-time control strategies for dynamical systems that involve frictional contact interactions. Hybridness and underactuation are key characteristics of these systems that  complicate the design of feedback controllers. In this research, we examine and test a novel feedback controller design on a planar pushing system, where the purpose is to control the motion of a sliding object on a flat surface using a point robotic pusher. The pusher-slider is a simple dynamical system that retains many of the challenges that are typical of robotic manipulation tasks.

  \hspace{3mm} Our results show that a model predictive control approach used in tandem with integer programming offers a powerful solution to capture the dynamic constraints associated with the friction cone as well as the hybrid nature of the contact. In order to achieve real-time control, simplifications are proposed to speed up the integer program. The concept of \textit{Family of Modes} (FOM) is introduced to solve an online convex optimization problem by selecting a set of contact mode schedules that spans a large set of dynamic behaviors that can occur during the prediction horizon. The controller design is applied to stabilize the motion of a sliding object about a nominal trajectory, and to re-plan its trajectory in real-time to follow a moving target. We validate the controller design through numerical simulations and experimental results on an industrial ABB IRB 120 robotic arm.
\end{abstract}
\section{INTRODUCTION}
Humans manipulate objects within their hands with impressive agility and ease. While doing so, they also make many and frequent mistakes from which they recover seamlessly. The mechanical complexity of the human hand along with its array of sensors sure play an important role. However, despite recent advances in the  design of complex robotic hands~\cite{ Kawasaki_2002, Mouri_2002, Gaiser_2008} and  sensory equipment (tactile sensors, vision markers, proximity sensors, etc.~\cite{Yousef_2011, Li_2014}), autonomous robotic manipulation remains far from human skill at manipulating with their hands or teleoperating robotic interfaces.

We argue that this gap in performance can largely be attributed to robots' inability to use sensor information for real-time control purposes. Whereas humans effectively process and react to information from tactile and vision sensing, robot manipulators are most often programmed in an open-loop fashion, incapable of adapting or correcting their motion. With the recent development of sensing equipment, the question remains:  how should robots use sensed information?

This work is concerned with the challenges involved in closing the loop through contact in robotic manipulation. To the knowledge of the authors, a general feedback controller design methodology is still lacking in the field of robotic manipulation, which is essential for robots to be aware and reactive to contact. In this article, we focus our attention on dexterous manipulation tasks, where the manipulated object moves relative to the robot's end effector. 

In this article we examine and test a feedback controller design for the pusher-slider system, where the purpose is to control the motion of a sliding object on a flat surface using a point pusher. The pusher-slider system is a simple dynamical system that incorporates several of the challenges that are typical of robotic manipulation tasks. In particular, we are concerned with two main challenges:
 \begin{enumerate}
 \item It is a \textbf{hybrid dynamical system} that exhibits different contact modes between the pusher and slider (e.g. separation, sticking, sliding up, and sliding down). Transitions between these modes result in discontinuities in the dynamics, which complicate  controller design.
 \item It is an \textbf{underactuated} system where the contact forces from the pusher acting on the sliding object are constrained to remain inside the friction cone. These constraints on the control inputs lead to a dynamical system where the velocity control of the pusher is not sufficient to produce an arbitrary acceleration of the slider. Ultimately, the controller must reason about finite horizon trajectories and not just instantaneous actuation.
 \end{enumerate}

The purpose of this article is to develop a feedback controller design that can handle both challenges described above. Our  results show that a model predictive control approach used in tandem with integer programming offers a powerful solution to capture the dynamic constraints associated with the friction cone as well as the hybrid nature of contact. In order to achieve real-time control, simplifications are proposed to speed up the integer program. The concept of \textit{Family of Modes} (FOM) is proposed to solve an online convex optimization problem by simulating the dynamical system forward using a set (i.e., family) of mode schedules that are identified as being key. Numerical simulations and experimental results, performed using an industrial robotic manipulator to push the sliding object and a Vicon system to track its pose, show that the FOM methodology yields a feedback controller design that can be implemented in real-time and stabilize the motion of a slider through a single contact point about a nominal trajectory. 
\section{RELATED WORK}
Historically, grasping and in-hand dexterous manipulation have been two main focuses of robotic manipulation research. In grasping, conventional control approaches first search for the location of grasp contact points that yield some form of geometric closure on the object~\cite{Bicchi_2000}, then close the gripper either blindly or with force control, and finally  treat the object as a rigid extension of the robotic arm. In-hand dexterous manipulation was first explored by~Salisbury and Craig~\cite{Salisbury_1982} for dynamic in-hand  motion. The controller design techniques presented in the in-hand dexterous literature typically apply to complex robotic hands and rely on the gripper and the manipulated objects to stick together at the contact points.

Dafle et al. \cite{Dafle_2014, Dafle_2015} demonstrated that simple robotic hands can be used to perform fast and effective regrasp strategies by using the robotic arm and the environment as an external source of actuation. When performing these regrasp strategies, the contact interactions are not limited to sticking contact but also necessarily exploit sliding motion. The control actions proposed by~\cite{Dafle_2014, Dafle_2015} require offline trajectory planning and rely on accurate contact models.

Recently, Posa et al.~\cite{Posa_2014} proposed to apply trajectory optimization tools to determine the motion of the robot and the manipulated object by including contact reaction forces along with motions as decision variables in a large optimization program. This method has been shown to be effective for path planning of high degree of freedom systems undergoing contact interactions.  This paper shares the motivation of including contact forces as decision variables as part of an optimization program. 

The application of feedback control strategies to discontinuous contact dynamical systems is a relatively unexplored field of research. Tassa and al.~\cite{Tassa_2010, Tassa_2012} have achieved remarkable simulation results by using smoothed contact models in an optimal control framework, and a similar approach has been proposed by Stewart and Anitescu~\cite{Stewart_2009}. These methods contrast to the approach explored in this paper, where hybridness of contact is considered explicitly.  

With regard to the pusher-slider system, Mason~\cite{Mason_1986} presented an early study of the mechanics of planar pushing. This theory has been applied to the design of controllers that  achieve stable pushing~\cite{Lynch_1996}, which offers the advantage of operating without sensor feedback by acting as an effective grasp. In~\cite{Lynch_1992}, this theory is expanded to a tactile feedback based controller for the case of a point pusher-slider system.

\section{CHALLENGES OF CONTROL THROUGH CONTACT}
The aim of this work is to accurately control the motion of objects through contact. Two major difficulties are typical of these systems: {hybridness and underactuation}.

\subsection{Hybridness}

When in contact, object and manipulator can interact in  different manners. For example, the object can slip within the fingers of the gripper, the gripper can throw the object in the air or perform pick and place maneuvers, etc. These manipulation actions correspond to different contact interaction modes, namely sliding, separation, and sticking. The hybridness associated with the transitions between modes can result in a non-smooth dynamical system. This complicates the design of feedback controllers as the vast majority of standard control techniques rely on smoothness of the dynamical model. 
 
\begin{figure}[h]
\centering
{	
		\includegraphics[width=12cm]{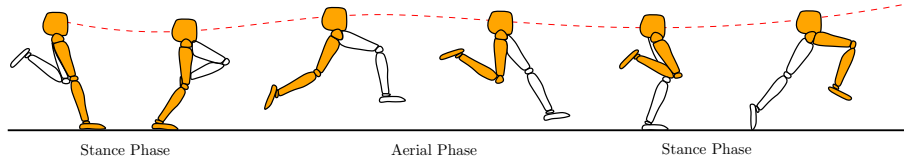}
}
\caption{\footnotesize{Human running gait adapted from Decker et al. \cite{Decker_2007}. The periodic nature of human gait permits to use control strategies that rely on offline mode scheduling.}} 
\label{SLIP_model}
\end{figure}

\begin{figure}[h]
\centering
{	
  \includegraphics[width=12cm]{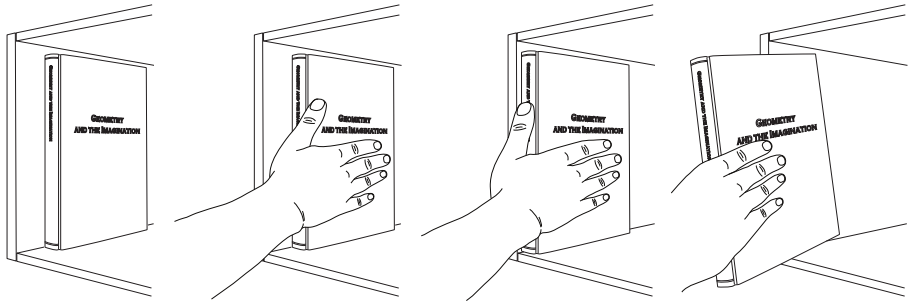}
}
\caption{\footnotesize{Animation of a simple manipulation task that exploits multiple contact modalities. First, the hand sticks to the book and drags it backwards exploiting friction. Second, thumb and fingers slide to perform a regrasp maneuver. Finally, the book is retrieved from the shelf using a stable grasp.}} 
\label{Manipulation_book}
\end{figure}

In many applications involving hybrid dynamical systems, this difficulty is overcome by setting the mode scheduling of the controller offline or using on-board sensing to detect mode transitions. For example, in the locomotion community, it is common to transition between two feedback controllers as the robot switches from a stance phase to an aerial phase~\cite{Poulakakis_2009}, as  in Fig.~\ref{SLIP_model}. 
For robotic manipulation tasks, the mode scheduling is often not known a priori and can be challenging to predict. In such cases, we must rely on the controller to decide during execution what interaction mode is most beneficial to the task. Figure~\ref{Manipulation_book} illustrates the example of picking a book from a shelf. The hand interacts with the book in a complex manner. It is difficult to say when fingers and palm stick or slide, but those transitions not only happen, but are necessary to pick the book. Likely the hand initially sticks to the book and drags it backwards exploiting friction. Then, the thumb and fingers swiftly slide to regrasp the book. Finally, the book is retrieved from the shelf using a stable grasp. For such manipulation tasks where the motion is not periodic, determining a fixed mode sequencing strategy is not obvious and likely impractical. Errors  during execution will surely require that the mode sequencing be altered.
%
\subsection{Underactuation}
Underactuation is due to the fact that contact interactions can only transmit a limited set of forces and torques to the object. As such, the controller must reason only among the forces that can physically be realized. For example, the normal forces commanded should be positive, as contact interactions can only ``push''  and cannot ``pull.'' In order to achieve this, it is required to explicitly integrate the physical constraints associated with contact interactions in the controller design.   A second important consequence of underactuation is that the controller must be capable of reasoning about future not just instantaneous actuation since the forces required to drive the task in the direction of the goal might not be feasible at the current instant. The controller must reason on a finite horizon.


\section{PUSHER-SLIDER SYSTEM}
\label{dynamic_model}
In this article, we study the pusher-slider system, a simple nonprehensile manipulation task where the goal is to control the motion of a sliding object (slider) through a single frictional contact point (pusher). The pusher-slider system is a useful test case dynamical system for controller design where  actuation arises from friction. 

\subsection{Kinematics}
\label{Dyn_Model}
\begin{wrapfigure}{r}{0.45\textwidth}
  \vspace{-15mm}
\centering
   \includegraphics[width=5cm]{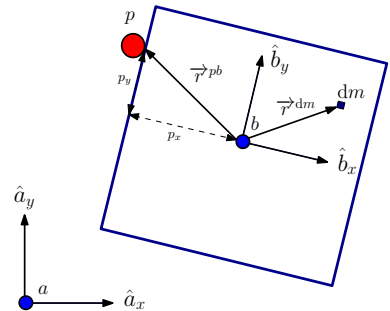}
\caption{\footnotesize{Kinematics of a slider  subject to  contact interactions with a single point of contact  robotic  pusher.}} 
  \vspace{-13mm}
\label{PusherSlider}
\end{wrapfigure}
Consider the  system  in Fig.~\ref{PusherSlider}. The pose of the slider is  
$
\mbf{q}_s = \bma{ccc}
x&y&\theta
\ema^\trans$ 
where $x$ and $y$ denote the cartesian coordinates of the center of mass  of the slider and $\theta$  its orientation relative to the inertial reference frame $\mathcal{F}_a$. The position of the pusher relative to point $b$ resolved in $\mathcal{F}_b$ is 
$
\mbf{r}_b^{pb} = \bma{cc}
p_x&p_y
\ema^\trans$. Figure~\ref{PusherSlider} shows the kinematics of the slider subject to  contact interactions with a single point of contact  robotic  pusher. 

\subsection{Quasi-Static Approximation}
During contact interactions, two external forces are exerted on the slider: the generalized frictional force applied by  the pusher on the slider (denoted $\mbf{f}^P$) and the generalized frictional force applied by the ground on the sliding object (denoted $\mbf{f}^G$). Applying Newton's second law in the $\hat{a}_x$ - $\hat{a}_y$ plane yields the motion equations
\beq
\mbf{H}\ddot{\mbf{q}}_s = \mbf{f}^G + \mbf{f}^P,
\label{DynamicEOM}
\eeq
where $\mbf{H}$ denotes the inertia matrix of the system.  The quasi-static assumption suggests that at low velocities,  frictional contact  forces dominate and   inertial forces do not have a decisive role in determining the motion of the slider. Under this assumption, the applied frictional force by the pusher is of equal magnitude and opposite direction to the ground planar frictional force (i.e., $ \mbf{f}^P = -\mbf{f}^G$). This quasi-static assumption leads to a simplified analysis of the motion of a sliding object using a single point of contact robotic pusher. Note that including the term $\mbf{H}\ddot{\mbf{q}}_s$ does not complicate the controller design and could easily be integrated into the control formulation presented in Section~\ref{MPC}. The resulting controller from a dynamic analysis  yields a mapping between the motion of the slider to the  reaction forces applied on the object. In contrast, the quasi-static assumption leads to a direct mapping between the motion of the slider and the motion of the pusher. This proved desirable from an experimental implementation standpoint using a position controlled robotic manipulator. 

The motion equations of the pusher-slider system are formulated  in \cite{Lynch_1992} assuming a quasi-static formulation with a uniform pressure distribution. Prior to presenting these motion equations, it is necessary to review two important concepts of frictional contact interactions: the limit surface and the motion cone.

\subsection{Limit Surface} 
The limit surface is a useful geometric representation which, under the quasi-static assumption, maps the applied frictional force on an object to its resulting velocity. First introduced in~\cite{Goyal_1991},  the limit surface is defined as a convex surface which bounds the set of all possible frictional forces and moments that can be sustained by frictional interface.
In this paper, we use the ellipsoidal approximation to the limit surface~\cite{Cutkosky_1991}, where the semi-principal axes are given by $f_{max}$, $f_{max}$, and $m_{max}$ defined by $f_{max} = \mu_g m g$ and 
$
m_{max} = \frac{\mu_g m g}{A}  \int_{\mathcal{A}} \norm{\ora{r}^{\mathrm{d}m\,b}} \mathrm{d}A,
$
where $\mu_g$ is the coefficient of friction between the object and the ground, $m$ is the mass of the object, $g$ is the gravitational acceleration, $A$ is the surface area of the object exposed to friction, and $\ora{r}^{\mathrm{d}m\,b}$ denotes the position of $\mathrm{d}m$ relative to the origin of $\mathcal{F}_b$. 

\subsection{Motion Cone} 
\label{MotionCone}
\begin{wrapfigure}{r}{0.35\textwidth}
\vspace{-10mm}
 \centering
   \includegraphics[width=4cm]{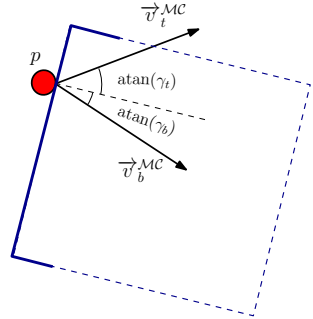}
\caption{\footnotesize{Motion cone ($\mathcal{MC}$) at contact point $p$. If the pusher velocity lies within the two boundaries of the motion cone, the pusher will stick to the slider, else, it will slide.}} 
\label{fig:MotionCone}
\vspace{-10mm}
\end{wrapfigure}
Depending on the direction of  motion of the pusher, different contact interaction modes can arise between the pusher and the slider. The motion cone~\cite{Mason_1986}, shown in Fig.~\ref{fig:MotionCone},   is useful to determine if a given velocity of the pusher will result in sticking or sliding behavior between the pusher and the slider.  Each boundary of the motion cone is constructed by mapping the resulting velocity of the slider at the contact point $p$  when subject to a frictional force that lies on a boundary of the friction cone. It can be shown using the ellipsoidal approximation to the limit surface, that for flat faced objects, the two boundaries of the motion cone are given as
$
\ora{v}^{\mathcal{MC}}_t = -1 \hat{b}_x + \gamma_t \hat{b}_y$
and
$
\ora{v}^{\mathcal{MC}}_b = -1 \hat{b}_x + \gamma_b \hat{b}_y,
$
with
%
%
\beq
\gamma_t= \frac{\mu c^2 - p_xp_y + \mu p_x^2}{c^2+p_y^2 - \mu p_x p_y}
\label{GammatEq}
\eeq
and
\beq
\gamma_b= \frac{-\mu c^2 - p_xp_y - \mu p_x^2}{c^2+p_y^2 + \mu p_x p_y},
\label{GammabEq}
\eeq
where $p_x$ and $p_y$ are shown in Fig.~\ref{PusherSlider} and $c=\frac{f_{max}}{m_{max}}$. Given the velocity of the pusher resolved in $\mathcal{F}_b$, denoted  $\mbf{u} = [v_n\,\,v_t]^\trans$, the conditions stated below determine the resulting contact interaction mode that will arise between the pusher and the slider.
 
\subsubsection{Sticking} 
When the sliding object is sticking to the pusher, the relative tangential velocity between the pusher and the object is zero. In order to have this behavior, the velocity vector must lie within the boundaries of the motion cone in Fig.~\ref{fig:MotionCone}. This constraint, denoted as $\mbf{u} \in \mathcal{MC}$, is defined  as
\begin{numcases}{\mbf{u}\in\mathcal{MC}:\hspace{2mm}}
v_t \leq \gamma_t v_n \label{Sticking_1}, \\ 
v_t \geq \gamma_b v_n \label{Sticking_2}. 
\end{numcases}
 \subsubsection{Sliding Up}
When the pusher is sliding up in the tangential direction relative  to the object, velocity of the pusher must lie above the upper boundary of the motion cone.  This constraint is expressed by
\begin{numcases}{\mbf{u}>\mathcal{MC}:\hspace{2mm}}
v_t > \gamma_t v_n \label{SlidingUp}.
\end{numcases}
 \subsubsection{Sliding Down}
When the pusher is sliding down in the tangential direction relative  to the object, the velocity of the pusher must lie below the lower boundary of the motion cone . This condition is enforced as
\begin{numcases}{\mbf{u}<\mathcal{MC}:\hspace{2mm}}
v_t < \gamma_b v_n \label{SlidingDown}.
\end{numcases}
\vspace{-10mm}

\subsection{Motion Equations}
\label{MotionEquations}
The motion equations of the pusher-slider system are formulated  in \cite{Lynch_1992} and stated below. The  equations presented in Eq.~\eqref{EOM} describe hybrid dynamics, where the contact interaction mode depends upon the direction of the pusher velocity. 
\beq
\mbfdot{x} =
\begin{cases}
 \mbf{f}_1(\mbf{x},\mbf{u}) \hspace{3mm} \text{if} \hspace{3mm} \mbf{u} \in \mathcal{MC},\\
 \mbf{f}_2(\mbf{x},\mbf{u})   \hspace{3mm} \text{if} \hspace{3mm} \mbf{u} > \mathcal{MC},\\
 \mbf{f}_3(\mbf{x},\mbf{u})   \hspace{3mm} \text{if} \hspace{3mm} \mbf{u} < \mathcal{MC},
 \end{cases}
 \label{EOM}
\eeq
with $\mbf{x} = [\mbf{q}_s^\trans \,\, p_y]^\trans$, $\mbf{u} = [v_n\,\,v_t]^\trans$ denotes the velocity of the pusher resolved in $\mathcal{F}_b$, and 
\bdis
\mbf{f}_j(\mbf{x}, \mbf{u}) =
\bma{ccc}
\mbf{C}^\trans \mbf{Q}\mbf{P}_j \\
\mbf{b}_j \\
\mbf{c}_j
\ema
\mbf{u},
\hspace{1mm}
\mbf{C} = \bma{cc}
\cos\theta & \sin \theta \\
-\sin \theta & \cos \theta
\ema
,\hspace{1mm}
\mbf{Q} = 
\frac{1}{c^2 + p_x^2 + p_y^2}
\bma{cc}
c^2 + p_x^2 & p_x p_y\\
 p_x p_y & c^2 + p_y^2
\ema,
\edis
\bdis
\mbf{b}_1 = 
\bma{ccc}
\frac{-p_y}{c^2 + p_x^2 + p_y^2} &\hspace{1mm} &p_x
\ema,
\hspace{2mm}
\mbf{b}_2 =
\bma{cc}
\frac{- p_y+\gamma_tp_x}{c^2 + p_x^2 + p_y^2}  & 0
\ema,
\hspace{2mm}
\mbf{b}_3 =
\bma{cc}
\frac{- p_y+\gamma_b p_x}{c^2 + p_x^2 + p_y^2} & 0
\ema,
\edis
\bdis
\mbf{c}_1 = \bma{cc}
0\\0
\ema^\trans
,
\hspace{1mm}
\mbf{c}_2 = \bma{cc}
-\gamma_t\\ 0
\ema^\trans
,
\hspace{1mm}
\mbf{c}_3 = \bma{ccc}
-\gamma_b\\0
\ema^\trans,
\hspace{1mm}
\mbf{P}_{1} = \mbf{I}_{2\times 2}
,
\hspace{1mm}
\mbf{P}_{2} = \bma{cc}
1 & 0 \\
\gamma_t & 0
\ema
,
\hspace{1mm}
\mbf{P}_{3} = \bma{cc}
1 & 0 \\
\gamma_b & 0
\ema,
\edis
where $j=1,2,3$ correspond to sticking, sliding up, and sliding down contact interaction modes, respectively. For simplicity, we do not consider the case of separation when the pusher is not in contact with the object. Under the assumption of small forces with low impact, separation is the least relevant mode to the pusher-slider system.

\subsection{Linearization}
This section develops the linearized motion equations and motion cone constraints developed in Sections~\ref{MotionCone} and~\ref{MotionEquations} about a given nominal trajectory. This linearization yields linear equations, which can be enforced as linear matrix inequalities in an optimization program and are computationally tractable for real-time execution of the controller design presented in Section~\ref{MPC}. Consider a feasible nominal trajectory $\mbf{x}^\star(t)$ of the sliding object with nominal control input $\mbf{u}^\star(t)$ of the pusher. The notation $(\cdot)^\star$ is used to  evaluate a term at the equilibrium state and $(\bar{\cdot})$ is used to denoted a perturbation about the equilibrium state. 
The linearization of motion equations Eq.~\eqref{EOM} about a nominal trajectory yields
\bdis
\dot{\mbfbar{x}} = \mbf{A}_j(t)\mbfbar{x} + \mbf{B}_j(t)\mbfbar{u},
\hspace{2mm}\text{with} \hspace{2mm} \mbfbar{x} = \mbf{x} - \mbf{x}^\star,
\hspace{2mm} \mbfbar{u} = \mbf{u} - \mbf{u}^\star,
\edis
and
\beq
\mbf{A}_j(t) = \left.\frac{\p \mbf{f}_j(\mbf{x}, \mbf{u})}{\p \mbf{x}}\right|_{\mbf{x}^\star(t), \mbf{u}^\star(t)},\hspace{2mm}
\mbf{B}_j(t) = \left.\frac{\p \mbf{f}_j(\mbf{x}, \mbf{u})}{\p \mbf{u}}\right|_{\mbf{x}^\star(t), \mbf{u}^\star(t)}.
\label{LinearDynamicsMatrices}
\eeq
%
Similarly, the  constraints enforcing a sticking interaction between the pusher and the slider presented in Eqs.~\eqref{Sticking_1} and~\eqref{Sticking_2} are perturbed about the nominal trajectory as
\beq
\left(v_t^\star + \bar{v}_t\right) \leq \left(\gamma_t^\star + \bar{\gamma}_t\right) \left(v_n^\star + \bar{v}_n\right), 
\hspace{2mm}\text{and}\hspace{2mm}
\left(v_t^\star + \bar{v}_t\right) \geq \left(\gamma_b^\star + \bar{\gamma}_b\right) \left(v_n^\star + \bar{v}_n\right),\label{StickingLinear}
\eeq
respectively. Expanding the perturbations $\bar{\gamma_t}$ and $\bar{\gamma_b}$  in terms of $\mbfbar{x}$ as
\beq
\bar{\gamma_t} = \mbf{C}_t \mbfbar{x}\hspace{2mm}, \hspace{2mm}
\bar{\gamma_b} = \mbf{C}_b\mbfbar{x},
\hspace{2mm} 
\mbf{C}_t=
\left.\frac{\p \gamma_t}{\p \mbfbar{x}}\right|_{\mbf{x}^\star(t),\mbf{u}^\star(t)},
\hspace{2mm}
\mbf{C}_b=
 \left.\frac{\p \gamma_b}{\p \mbfbar{x}}\right|_{\mbf{x}^\star(t),\mbf{u}^\star(t)},
 \label{Cs}
\eeq 
permits to write Eq.~\eqref{StickingLinear} in matrix form as 
\beq
\mbf{E}_1(t)\mbfbar{x}+\mbf{D}_{1}(t) \mbfbar{u}\leq \mbf{g}_1(t),
\label{StickingMatrixLinearized}
\eeq 
where $\mbf{E}_1 = v_n^\star\bma{cc}
-\mbf{C}_t \\
\mbf{C}_b
\ema$, 
$\mbf{D}_1 = 
\bma{cc}
-\gamma_t^\star & 1\\
\gamma_b^\star & -1
\ema$, $\mbf{g}_1 = \bma{cc}
-v_t^\star + \gamma_t^\star v_n^\star\\
v_t^\star - \gamma_b^\star v_n^\star
\ema$, where higher order perturbations are neglected. Similarly, the sliding up and sliding down constraints in Eqs.~\eqref{SlidingUp} and~\eqref{SlidingDown}   are perturbed about the nominal trajectory as
\beq
\left(v_t^\star + \bar{v}_t\right)  >  \left(\gamma_t^\star + \bar{\gamma}_t\right) \left(v_n^\star + \bar{v}_n\right), \hspace{2mm}\text{and}\hspace{2mm}
\left(v_t^\star + \bar{v}_t\right)  <  \left(\gamma_b^\star + \bar{\gamma}_b\right) \left(v_n^\star + \bar{v}_n\right),
\label{Sliding}
\eeq
respectively and can be rearranged in matrix form as 
\beqarray
\mbf{E}_2(t)\mbfbar{x}+\mbf{D}_{2}(t) \mbfbar{u}&\leq& \mbf{g}_2(t),
\label{SlidingUpMatrixLinearized}
\hspace{2mm}\text{and}\hspace{2mm}
\\
\mbf{E}_3(t)\mbfbar{x}+\mbf{D}_{3}(t) \mbfbar{u}&\leq& \mbf{g}_3(t),
\label{SlidingDownMatrixLinearized}
\eeqarray
with $\mbf{E}_2 = v_n^\star
\mbf{C}_t $,  $\mbf{E}_3 = -v_n^\star
\mbf{C}_b $, 
$\mbf{D}_2 = 
\bma{cc}
\gamma_t^\star & -1
\ema$, $\mbf{D}_3 = 
\bma{cc}
-\gamma_b^\star & 1
\ema$,  $\mbf{g}_2 = \bma{cc}
v_t^\star - \gamma_t^\star v_n^\star - \epsilon
\ema$, and $\mbf{g}_3 = \bma{cc}
-v_t^\star + \gamma_b^\star v_n^\star - \epsilon
\ema$, where higher order perturbations are neglected, $\epsilon$ is a small scalar value, and $\mbf{C}_t$ and $\mbf{C}_b$ are given by Eq.~\eqref{Cs}. 
\section{MODEL PREDICTIVE CONTROL}
\label{MPC}

\begin{figure}[h]
\centering
{	
	\includegraphics[width=10cm]{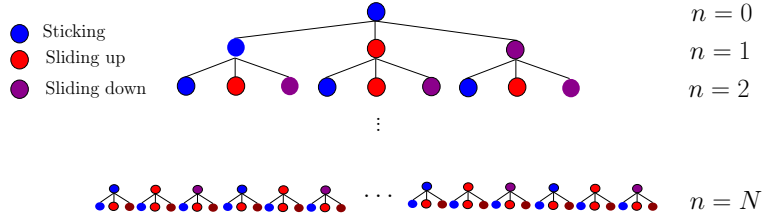}

\caption{\footnotesize{Tree of optimization programs for a MPC program with $N$ prediction steps. Scales exponentially due to contact hybridness.}} 
\label{MPC_Tree}	
}
\end{figure}

In this section, we present a feedback controller design for the motion of the slider that minimizes perturbations from a desired trajectory. 
The proposed controller determines the desired pusher velocity  at each time step based on the sensed pose of the slider. A successful feedback controller must: 1)  address  hybridness and underactuation 2) allow for sliding at contact 3) be  fast enough to solve online 4) drive perturbations from the nominal trajectory to zero. 
To satisfy these requirements, we use a Model Predictive Control (MPC) formulation, which takes the form of an optimization program over the control inputs during a finite time horizon  ${t}_0, \hdots, {t}_N$.
The decision variables of the optimization program include the perturbed states of the system for $N$ time steps $\mbfbar{x}_1, \hdots, \mbfbar{x}_N$ and the perturbed control inputs $\mbfbar{u}_0, \hdots, \mbfbar{u}_{N-1}$. 
The goal is represented by a finite-horizon cost-to-go function that we will minimize subject to the constraints on the control inputs and the dynamics of the system detailed in Section~\ref{dynamic_model}. 
We express the cost-to-go for $N$ time steps as:
\beq
J(\mbfbar{x}_n,\mbfbar{u}_n) = \mbfbar{x}_N^\trans\mbf{Q}_N\mbfbar{x}_N + \sum_{n=0}^{N-1} \left(\mbfbar{x}_{n+1}^\trans\mbf{Q}\mbfbar{x}_{n+1} + \mbfbar{u}_n^\trans \mbf{R}\mbfbar{u}_n\right).
\label{opt_prog}
\eeq
The terms $\mbf{Q}$, $\mbf{Q}_N$, and $\mbf{R}$ denote weights matrices associated with the error state, final error state, and control input, respectively. We subject the search for optimal control inputs to the constraints describing the motion equations and contact dynamics of the system. Due to the hybridness of the dynamical  equations, the constraints to be enforced depend on the contact mode $j$ at play at each iteration $n$ of the prediction finite horizon:

\begin{numcases}{\hspace{-16mm}\text{if }n=0:\hspace{2mm}}
 \mbfbar{x}_1 =\mbf{f}_{j0}+ h\mbf{B}_{j0}\mbfbar{u}_0  \label{dynamics_x0}, & \hspace{16mm} \text{(Dynamics)} \\
 \mbf{D}_{j0}\mbfbar{u}_0 \leq\mbf{g}_{j0} \label{MC_02}, &\hspace{16mm} \text{(Motion Cone)}
 \end{numcases}
\begin{numcases}{\text{if }n>0:\hspace{2mm}}
{\mbfbar{x}}_{n+1} =\left[\mbf{I} + h \mbf{A}_{jn} \right]{\mbfbar{x}}_{n}+ h \mbf{B}_{jn}{\mbfbar{u}}_{n}, \label{dynamics_xn} &
\hspace{-3mm}\text{(Linearized Dynamics)} \\
\mbf{E}_{jn}\mbfbar{x}_n+\mbf{D}_{jn}\mbfbar{u}_n \leq\mbf{g}_{jn},  \label{MC_n} 
&\hspace{-3mm}\text{(Linearized Motion Cone)}
\end{numcases}
%
where the terms $\mbf{A}_{jn}$, $\mbf{B}_{jn}$, $\mbf{D}_{jn}$, $\mbf{E}_{jn}$ are developed in Eqs.~\eqref{LinearDynamicsMatrices}, \eqref{StickingMatrixLinearized}, \eqref{SlidingUpMatrixLinearized}, and \eqref{SlidingDownMatrixLinearized}, and the subscript $n$ is used to denote the time stamp at which each expression is evaluated (e.g. $\mbf{A}_{jn} = \mbf{A}_j(t_n)$).
The constraints Eqns.~\eqref{dynamics_x0} and~\eqref{dynamics_xn} describe the dynamical motion equations while  Eqns.~\eqref{MC_02} and~\eqref{MC_n} represent the motion cone constraints. Note that for the special case ($n=0$), the  nonlinear dynamical  and contact constraint equations given by Eqs.~\eqref{Sticking_1},~\eqref{Sticking_2},~\eqref{SlidingUp}, ~\eqref{SlidingDown}, and \eqref{EOM} reduce to linear form  due to the knowledge of the state $\mbf{x}_0$. In this case, the expressions $\mbf{B}_{j0}$, $\mbf{D}_{j0}$, and $\mbf{g}_{j0}$ are evaluated at $\mbf{x}_0$ rather than $\mbf{x}^\star (t_0)$,  and  the term $\mbf{f}_{j0}$ is defined as
\beqarraynn
\mbf{f}_{j0} &=& \mbfbar{x}_0 + h\left[\mbf{B}_{j0}\mbf{u}^\star(t_0) - \mbf{f}(\mbf{x}^\star(t_0), \mbf{u}^\star(t_0))\right].
\eeqarraynn
The constraints in Eqs. \eqref{dynamics_x0}, \eqref{MC_02}, and \eqref{dynamics_xn},~and \eqref{MC_n} depend on the contact mode $i$, which complicates the search for optimal control inputs. Contact modes and control inputs must be chosen simultaneously. As illustrated in Fig.~\ref{MPC_Tree}, this problem takes the form of a tree of optimization programs with $3^N$ possible contact schedules, each yielding a convex optimization program, which is too computationally expensive to solve online.
\vspace{-3mm}

\subsection{Mixed-Integer Quadratic Program}
\label{MIQP}

The combinatorial hybrid nature of the pusher-slider dynamics can be modeled by adding integer decision variables into the optimization program, as is commonly done in Mixed-Integer programming. The resulting Mixed-Integer Quadratic Program (MIQP) can be  solved rather efficiently using numerical tools, such as Gurobi~\cite{gurobi_2015}. In the case of the pusher-slider system, we introduce the integer variables:
$
z_{1n} \in \{0,1 \}$,
$z_{2n} \in \{0,1 \}$, and 
$z_{3n} \in \{0,1 \},
$
where $z_{1n}=1$, $z_{2n}=1$, or $z_{3n}=1$ indicate that the contact interaction mode at step $n$ is either sticking, sliding up, or sliding down, respectively. We will use the big-M formulation \cite{book_Nemhauser} to write down the problem, where $M$ is a large scalar value used to activate and deactivate the contact mode dependent constraints, through a set of linear equations.  The mode dependent constraints are reformulated as
\bdis
\text{if $n=0$: }
\begin{cases}
\bma{cc}
1\\-1
\ema
\mbfbar{x}_1 \leq 
\bma{cc}
1\\-1
\ema
\left\{
\mbf{f}_{j0}+ h\mbf{B}_{j0}\mbfbar{u}_0\right\}
+
\mbf{1}_{8\times 1} M(1-z_{jn})
\\
 \mbf{D}_{j0}\mbfbar{u}_0 \leq\mbf{g}_{j0} \label{MC_0}  + \mbf{1}_{2\times 1} M(1-z_{jn})
\end{cases}
\hspace{30mm}
\edis
\bdis
\text{if $n>0$: }
\begin{cases}
\bma{cc}
1\\-1
\ema
{\mbfbar{x}}_{n+1} \leq 
\bma{cc}
1\\-1
\ema
\left\{
\left[\mbf{1} + h \mbf{A}_j \right]{\mbfbar{x}}_{n}+ h \mbf{B}_j{\mbfbar{u}}_{n} \right\}
+
\mbf{1}_{8\times 1} M(1-z_{jn})
\\
\mbf{E}_{jn}\mbfbar{x}_n+\mbf{D}_{jn}\mbfbar{u}_n \leq\mbf{g}_{jn}  + \mbf{1}_{2\times 1} M(1-z_{jn}),
\end{cases}
\hspace{3mm}
\edis
where $\mbf{1}_{m\times1} = [1\,\, 1\,\, \hdots \,\,1]^\trans$.
Finally, the constraint $z_{1n}+z_{2n}+z_{3n} = 1$ is enforced to ensure that only one mode can be activated at a time. 
\subsection{Family of Modes (FOM) Scheduling}
\label{FOM}
\begin{figure}[t]
\centering
{	
		\includegraphics[width=10cm]{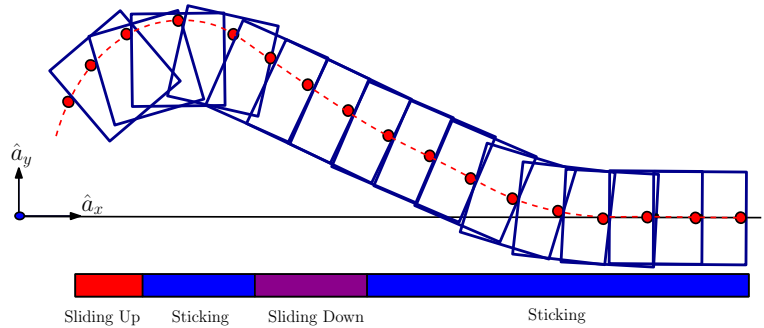}

\caption{\footnotesize{Example of optimal mode schedule for the pusher-slider system converging to a straight horizontal trajectory.}} 
\label{FOM_intro}	
}
\end{figure}

The MIQP formulation greatly reduces the computational cost associated with the optimization program in~Eq.~\eqref{opt_prog}. With an efficient implementation, it can be solved in almost real-time for the low dimensional pusher-slider system. However it does not scale well for systems with more degrees of freedom or additional contact points.  The method presented in this section is motivated by the observation that many of the branches of the tree in Fig. \ref{MPC_Tree} will give very good solutions, even if not exactly optimal. For the pusher-slider system, it is reasonable to expect the optimal mode schedule will follow a certain predictable structure. For example, 
%
%
we can intuitively expect that when the object is located in the $y$ direction above the reference straight line trajectory, as  in Fig. \ref{FOM_intro}, the pusher will likely slide up to correct the orientation of the object followed by a downward sliding motion as the object converges to the desired trajectory. This pushing strategy represents one possible mode schedule. 
%
The family of modes algorithm consists in determining a fixed set of probable mode schedules that span a large range of primitive behaviors of the system. Each mode schedule in the family specifies a sequence of $N$ contact modes to be imposed during the finite prediction horizon in the MPC formulation. By doing so, the combinatorial problem reduces to solving $m$ convex optimization programs, where $m$ is the number of mode sequences in the family.  A key challenge is in determining a small number of mode schedules that spans a ``significant'' set of dynamic behaviors. 
For the pusher-slider system, one could consider a family of three mode sequences:
%
%
\begin{itemize}
\renewcommand\labelitemi{}
	\item $\mathcal{M}_1$: the pusher slides up relative to the object followed  by a sticking phase,
	\renewcommand\labelitemi{}
	\item $\mathcal{M}_2$: the pusher  slides down relative to the object followed  by a sticking phase,
	\renewcommand\labelitemi{}
	\item $\mathcal{M}_3$: the pusher sticks to the object for the full length of the prediction horizon.
\end{itemize}
Even though this family of mode sequences only contains a very small fraction of all the possible contact mode combinations in the tree in Fig. \ref{MPC_Tree}, it spans a very large set of dynamic behaviors between the pusher and the slider. Part of the reason is that the controller will re-optimize the selection of optimal modes in real-time. Solving Eq. \eqref{opt_prog} for each mode schedule leads to the finite horizon costs $J_1$, $\hdots$, $J_m$. Given that all possible contact modes are predetermined, all combinatorial aspects disappear, and each mode schedule in the family leads to a computationally solvable quadratic problem. The controller then chooses the optimal among the ``m'' mode schedules. The control input is selected  at each time step by choosing the first element of the sequence of control inputs as $\mbf{u} = \mbf{u}_0^\star + \mbfbar{u}_0$, where  the term $\mbfbar{u}_0$ is obtained from the optimization program with minimum cost and $\mbf{u}^\star$ denotes the nominal control input.
%
\begin{table}[]
\vspace{-5mm}
\caption{Physical parameters of pusher-slider system.}
\vspace{-5mm}
\begin{center}
\label{table:table1}
\begin{tabular}{l l l l l}
& &  \\ 
 Property & Symbol  & Value\\
\hline
coefficient of friction (pusher-slider) & $\mu_p$ &  $ 0.3 $ \\
coefficient of friction (slider-table) & $\mu_g$ &  $ 0.35 $ \\
mass of slider, $kg$ & m & 1.05 \\
length of slider, $m$ & a & 0.09 \\
width of slider, $m$ & b & 0.09
\end{tabular}
\end{center}
\vspace{-10mm}
\end{table}
\section{RESULTS}
\label{RESULTS}
\begin{wrapfigure}{h}{0.35\textwidth}
\vspace{-8mm}
\centering
   \includegraphics[width=5.cm]{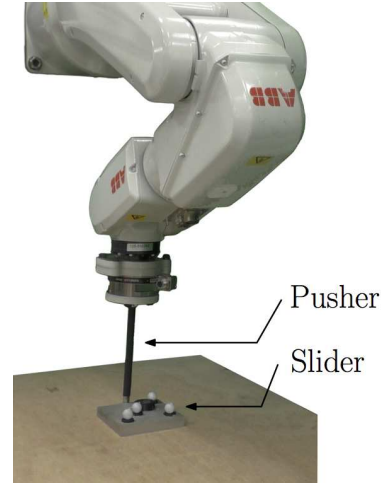}
\caption{\footnotesize{Experimental setup. A metallic rod (pusher) is attached to an ABB IRB 120 industrial robotic arm to push an aluminum object (slider). The pose of the slider is tracked using a Vicon camera system.}} 
\label{Robot}
\vspace{-5mm}
\end{wrapfigure}
The controller design based on the Family of Modes approach is implemented in this section as described in Section~\ref{MPC}. Two test scenarios are considered. Section~\ref{sec:StraightLineTracking} considers a trajectory tracking problem, where an external perturbation is applied to the system. Section~\ref{sec:TargetTracking} adapts the controller design to a target tracking problem, where the pusher  guides the slider through  $3$ successive targets. We evaluate the performance of the controller design in both test scenarios through numerical simulations and experiments. The experiments are conducted using an ABB IRB 120 industrial robotic manipulator along with  a Vicon system to track the pose of the slider. The experimental setup is depicted in Fig.~\ref{Robot}, where a metallic rod (pusher) attached to the robot is used to push an aluminum object (slider) on a flat surface (plywood). The physical parameters of the system are reported in Table~\ref{table:table1}.

\subsection{Straight Line Tracking}
\label{sec:StraightLineTracking}
\begin{figure}[h]
\vspace{-3mm}
\centering
\subfigure[Simulation results.]
{	
		\includegraphics[width=8cm]{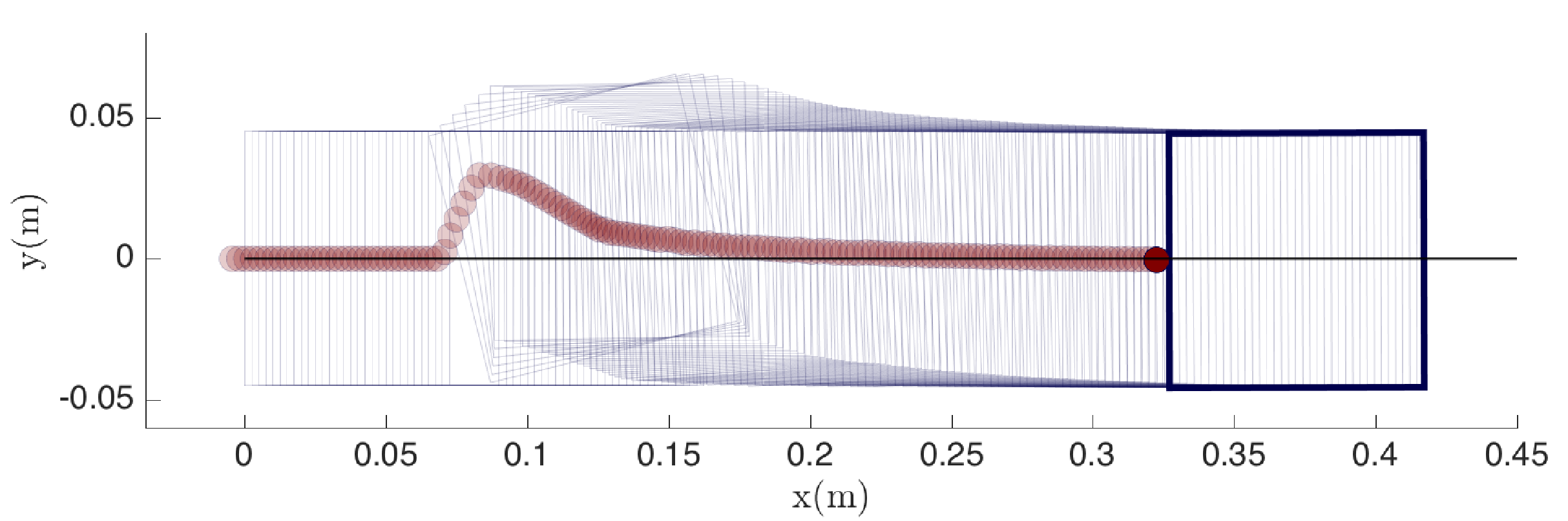}
		\label{SimStraightLinePlot}
}
\vspace{-3mm}
\subfigure[Experimental results.]
{
		\includegraphics[width=8cm]{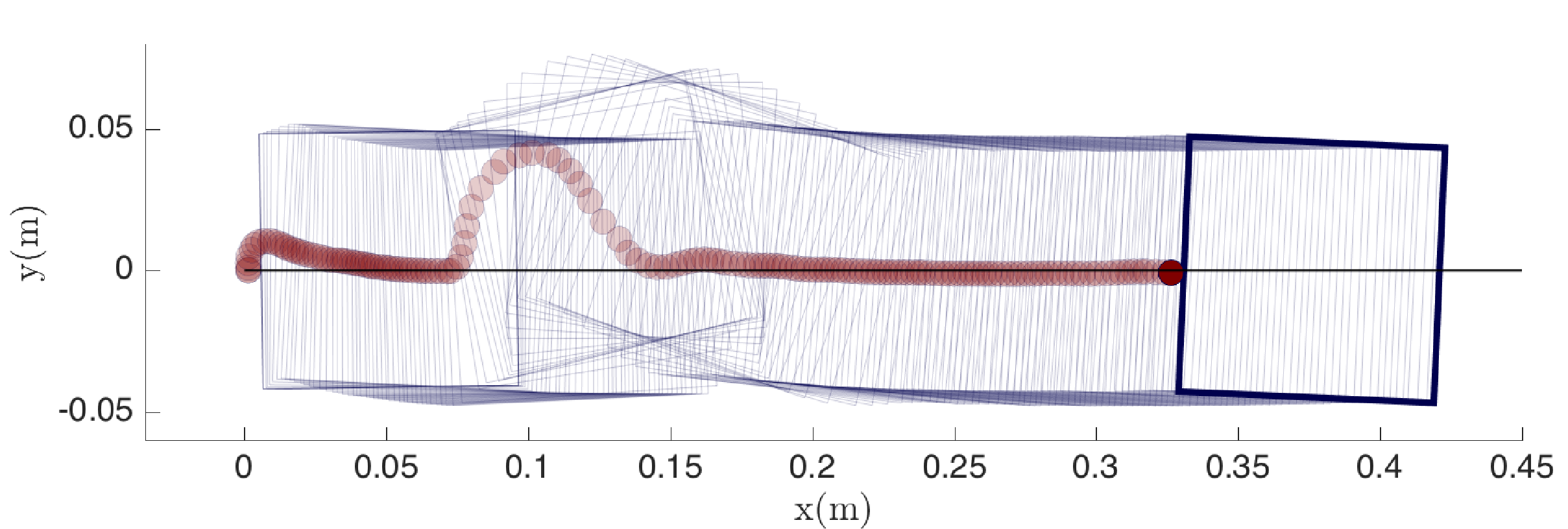} %
		\label{ExpStraightLinePlot}
}
\centering
\caption{\footnotesize{Tracking of a straight line trajectory at a constant velocity. Both simulated and experimental trajectories of the sliding object recover from an external lateral perturbation. The pusher quickly reacts to stabilize the slider about the nominal trajectory.}} 
\label{StraightLineTracking}
\end{figure}

Consider the problem of tracking a straight line nominal trajectory at a constant velocity, defined by 
$
\mbf{x}^\star(t) = [0.05 t\,\,0\,\,0\,\,0]^\trans $ and $
\mbf{u}^\star  = [0.05\,\,0]^\trans.
$
\bdis
\mathcal{M}_1 := \begin{cases}
\text{Slide up} \hspace{3mm} \text{if } n=0 \\
\text{Stick} \hspace{8mm} \text{if }  n > 0 
\end{cases}
\hspace{-3mm},\hspace{2mm}
\mathcal{M}_2 := \begin{cases}
\text{Slide down} \hspace{3mm} \text{if } n=0 \\
\text{Stick}\hspace{12mm} \text{if }  n > 0
\end{cases}
\hspace{-3mm},\hspace{2mm}
\mathcal{M}_3 := 
\text{Stick},\hspace{33mm}
\edis
as detailed in Section \ref{FOM}.  The controller design parameters used in the numerical simulations  in Fig.~\ref{SimStraightLinePlot.ps} and the experiments in Fig.~\ref{ExpStraightLinePlot} are  set to $N=35$ steps, $h=0.03$ seconds,  $\mbf{Q} = 10~\text{diag}\{1,3,.1,0\}$, $200~\text{diag}\{1,3,.1,0\}$, and $\mbf{R} = 0.5~\text{diag}\{1,1\}$.  To limit the maximum velocity of the pusher, we include the constraints $
\abs{v_n} \leq 0.1$ m/s and $
\abs{v_t} \leq 0.1$ m/s to the optimization program. 

%
In Fig.~\ref{StraightLineTracking}, an impulsive force is applied in the $y$ direction to perturb the system about its nominal trajectory and evaluate the performance of the feedback controller. In Fig.~\ref{SimStraightLinePlot}, the simulated response of the slider to a state perturbation $\mbs{\delta} = [0\,\,0.01\,\, \frac{15\pi}{180}\,\,0]^\trans$ applied at $x=0.075$ m  is compared to the experimental response of the slider in Fig.~\ref{ExpStraightLinePlot} to an external impulsive force applied using a hand held poker. Both simulated and experimental responses show that the feedback controller is successful in  driving the perturbations from the nominal trajectory to zero. The initial motion of the pusher in Fig.~\ref{ExpStraightLinePlot} is trying to correct the non-zero initial conditions of the slider, which was imperfectly placed by hand. The experimental performance of the feedback controller design to a variety of external perturbations can be visualized at  \url{https://mcube.mit.edu/videos}.
\subsection{Trajectory Tracking}
\label{sec:TargetTracking}
\begin{figure}
\vspace{-3mm}
\centering
\begin{minipage}{.4\textwidth}
  \centering
  \includegraphics[width=1.1\linewidth]{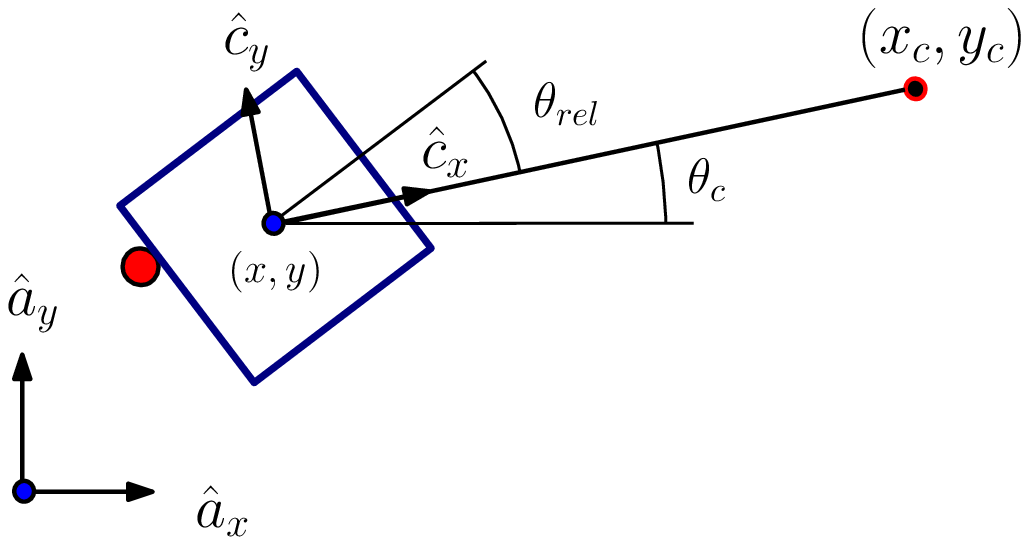}
  \captionof{figure}{\footnotesize{The intermediate reference frame $\mathcal{F}_c$ is defined such that $\hat{c}_x$ points in the direction of the target.}}
  \label{slider_blue}
\end{minipage}%
\hspace{8mm}
\begin{minipage}{.4\textwidth}
  \centering
  \includegraphics[width=1.1\linewidth]{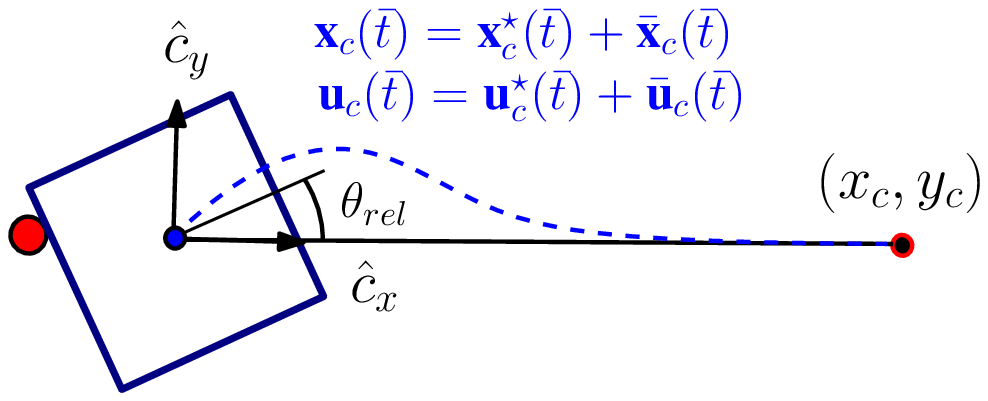}
  \captionof{figure}{\footnotesize{A desired trajectory $\mbf{x}_c^\star(\bar{t})$ is defined from the center of the object to the goal position along the direction $\hat{c}_x$ with nominal control input sequence $\mbf{u}_c^\star(\bar{t})$.}}
  \label{slider_blue_rotated}
\end{minipage}
\vspace{-5mm}
\end{figure}

\begin{figure}[h]
\vspace{-5mm}
\centering
\subfigure[Simulation results.]
{	
		\includegraphics[width=5.9cm]{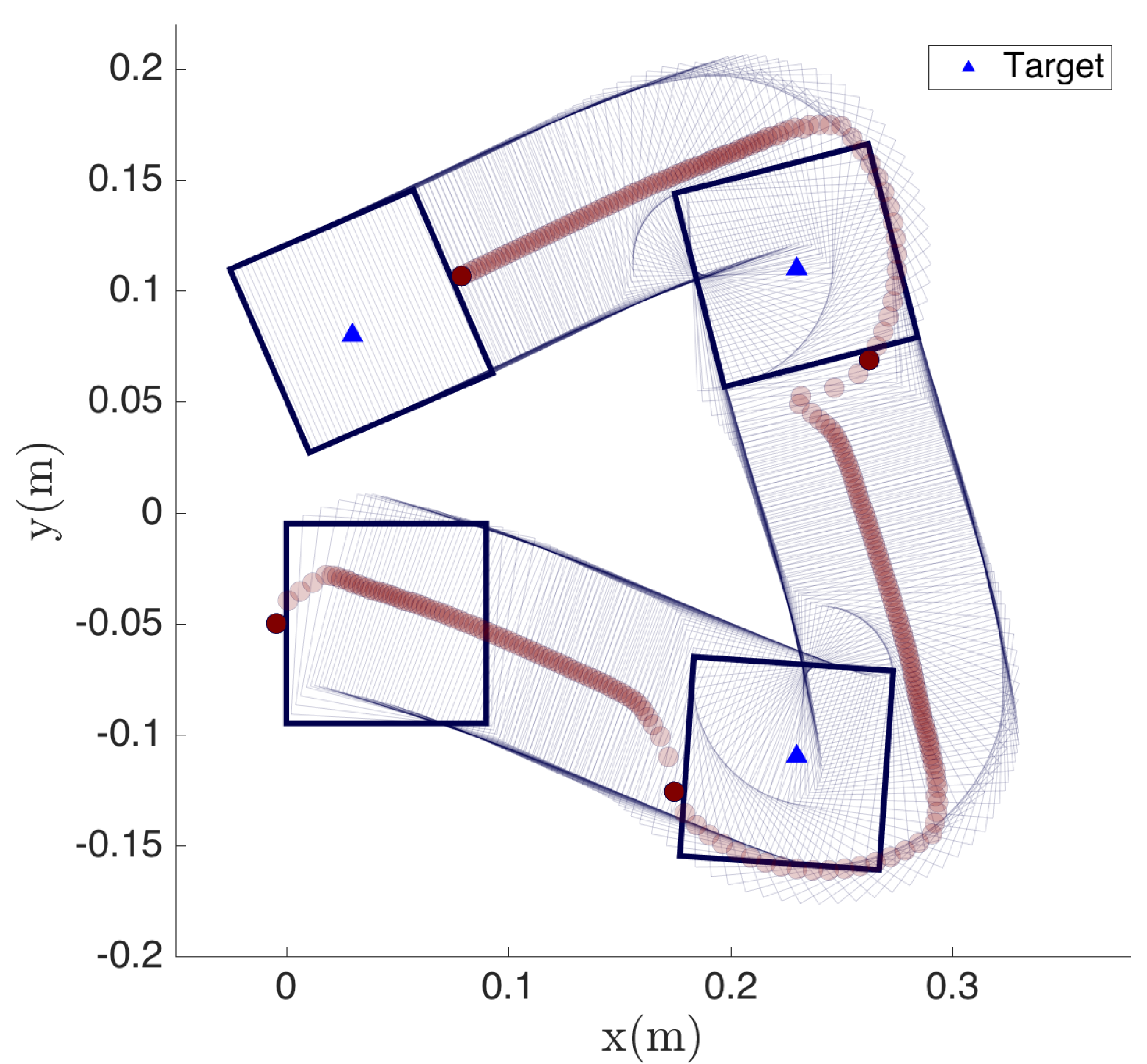}
		\label{SimTargets}
}
\hspace{-2mm}
\subfigure[Experimental results.]
{
		\includegraphics[width=5.9cm]{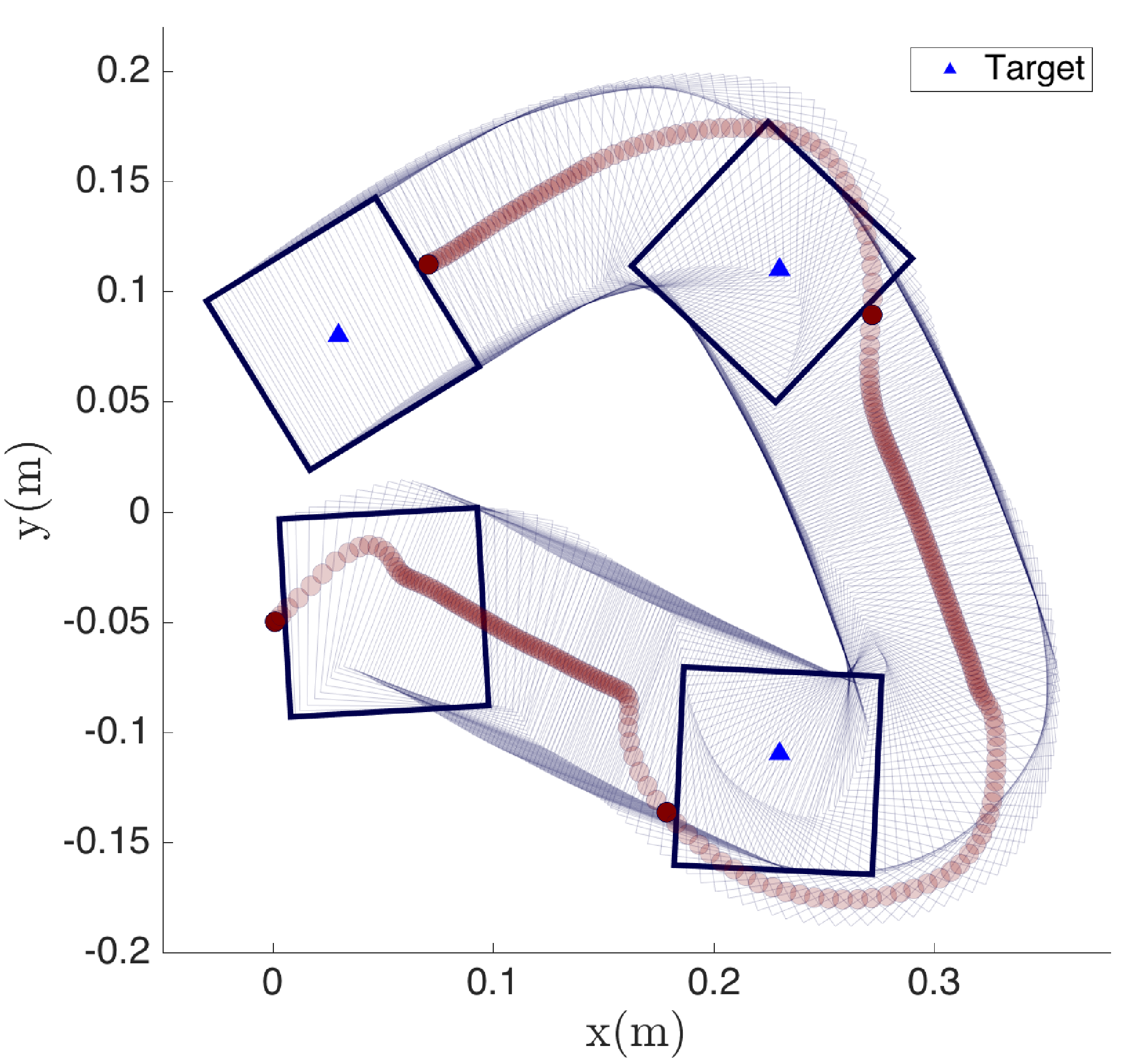} %
		\label{ExpTargets}
}

\centering
\caption{\footnotesize{Control of a sliding object through a single point robotic pusher.  Both simulation and experimental results successfully reach all $3$ targets within the specified tolerance.}} 
\label{TargetTracking}
\vspace{-4mm}
\end{figure}

In this section, the controller design developed in Sections~\ref{MPC} and \ref{RESULTS} is adapted  to the problem of tracking a moving target position. The objective is to control the motion of the robotic pusher such that the sliding object reaches a target ($x_c$, $y_c$).
At each instant,  an intermediate reference frame $\mathcal{F}_c$ is defined where the unit vector $\hat{c}_x$ points from the center of mass of the sliding object to the target position. 
The angle $\theta_c$ in Fig. \ref{slider_blue} is the orientation of $\hat{c}_x$ relative to the horizontal $\hat{a}_x$ and the angle $\theta_{rel} = \theta-\theta_c$ is the orientation of the slider relative to $\hat{c}_x$.
%
%
%
%
%
The tracking  of a target position is converted into a trajectory tracking problem where the objective is to track the straight line between the current position of the slider and the target. Assuming a constant desired velocity of the sliding object, the nominal trajectory $\mbf{x}_c^\star(\bar{t})$ and control input $\mbf{u}_c^\star(\bar{t})$, defined relative to the intermediate reference frame $\mathcal{F}_c$, are  
$
\mbf{x}_c^\star(\bar{t}) = \bma{cccc}
v_x \bar{t}&0&0&0 
\ema^\trans
$
and
$
\mbf{u}_c^\star(\bar{t}) = \bma{cc}
v_x&0
\ema^\trans,
$ 
with $v_x$  the desired velocity. The term $\bar{t}$ denotes the prediction horizon time, which is reinitialized at each time step as $\bar{t} = 0$.
%
%
The controller design parameters used in Fig.~\ref{TargetTracking}  are identical to those presented in Section~\ref{StraightLineTracking}, with the exception that the max tangential velocity is set to $
\abs{v_t} \leq 0.3$ m/s.  The target tracking results are performed from zero initial conditions with the target positions $[x_c\,\,y_c]^\trans$ given by 
\bdis
%
\text{Target 1:\,\,}\bma{cc}0.23\\-0.11\ema (m), \hspace{2mm}
\text{Target 2:\,\,}\bma{cc}0.23\\0.11\ema (m), \hspace{2mm}
\text{Target 3:\,\,}\bma{cc}03\\0.08\ema (m)
\edis
%
The simulation begins with the position tracking of Target $1$. When the slider falls within a distance of $0.01$ meters of a target, the position is updated to the next target position and so on until the final position is reached.  Both simulated and experimental trajectories  achieve  target tracking within the specified tolerance (video of experiments available at \url{https://mcube.mit.edu/videos}). 
 %
The trajectories in Fig. \ref{TargetTracking} depict the robotic pusher favoring a sliding behavior when the relative angle of the sliding object is large relative to the target position. The controller elects to slide the pusher relative to the object to rotate it and then favors a sticking contact mode to push the object in a straight line towards the target position. This control strategy is intuitive and is in line with the way in which humans manipulate objects using a single finger. It is observed that the simulated trajectory results in Fig.~\ref{SimTargets} achieves slightly more aggressive turns than the experimental trajectory results in Fig.~\ref{ExpTargets}. 
Despite the unmodeled aspects of the problem, such as delay in the robot position control, quasi-static assumption, imprecise nature of friction, etc.,   the feedback controller makes good control decisions which drive the system in the right direction towards the desired trajectory.
%
%
%
%
%
\section{CONCLUSION}
In this work we present a feedback controller design for manipulation tasks where control acts through contact. Using a model predictive control approach combined with an integer programming formulation permits for the integration of the physical constraints associated with the dynamics of contact as well as explicitly consider the hybrid nature of the interactions. We describe two methodologies to solve the optimization program, namely Mixed-Integer Quadratic Programming and Family of Modes, which we propose to speed up the integer program for real-time control purposes. The Family of Modes \ method is validated through numerical simulations and experiments, where the feedback controller successfully stabilizes  the motion of a sliding object through single point contact.  

Following the success of the Family of Modes approach in the pusher-slider system, a natural question raised is: ``How does the Family of Modes approach extend to more complex  manipulation tasks?'' For large set of manipulation problems involving multiple contact locations, key mode sequences can be readily identified either using physical insight of by investigating solutions to offline planning algorithms. Moreover, for many manipulation tasks, such as those involving parallel jaw grippers, there is a natural symmetry of the problem that can be leveraged to deduce important mode sequences.  The application of the Family of Modes approach to multiple contact problems, such as in-hand manipulation with extrinsic dexterity~\cite{Dafle_2014,Dafle_2015}, is a  direction of future work.  
\addcontentsline{toc}{section}{References}
\bibliographystyle{aiaa}
\bibliography{wafr_bib}

\end{document}